\documentclass{article}

\PassOptionsToPackage{numbers, compress}{natbib}

\usepackage[preprint]{neurips_2020}

\usepackage{hyperref}       
\usepackage{url}            
\usepackage{booktabs}       
\usepackage{amsfonts}       
\usepackage{nicefrac}       
\usepackage{microtype}      
\hypersetup{
    colorlinks=true,
    citecolor=black,
    urlcolor=blue,
}
\usepackage{amsfonts}       
\usepackage{amsmath, amsthm, amssymb}
\usepackage{bm}
\usepackage{graphicx}
\usepackage{multirow}
\usepackage{tabularx}
\usepackage{booktabs}
\usepackage[ruled,vlined]{algorithm2e}
\usepackage{algorithmic}
\let\vec\mathbf

\newtheorem{theorem}{Theorem}[section]

\newtheorem{lemma}[theorem]{Lemma}
\newtheorem{definition}[theorem]{Definition}

\usepackage{wrapfig}

\title{Interval Deep Learning for Uncertainty Quantification in Safety Applications}

\author{%
  David Betancourt\\
  School of Computational Science and Engg\\
  Georgia Institute of Technology\\
  Atlanta, GA 30332\\
  \texttt{david.betancourt@gatech.edu} \\
   \And
   Rafi L. Muhanna \\
   School of Civil and Environmental Engg \\
   Georgia Institute of Technology\\
   Atlanta, GA 30332\\
   \texttt{rafi.muhanna@gatech.edu} \\
}

\begin{document}

\maketitle

\begin{abstract}
    Deep neural networks (DNNs) are becoming more prevalent in important safety-critical applications, where reliability in the prediction is paramount. Despite their exceptional prediction capabilities, current DNNs do not have an implicit mechanism to quantify and propagate significant input data uncertainty---which is common in safety-critical applications. In many cases, this uncertainty is epistemic and can arise from multiple sources, such as lack of knowledge about the data generating process, imprecision, ignorance, and poor understanding of physics phenomena. Recent approaches have focused on quantifying parameter uncertainty, but approaches to end-to-end training of DNNs with epistemic input data uncertainty are more limited and largely problem-specific. In this work, we present a DNN optimized with gradient-based methods capable to quantify input and parameter uncertainty by means of interval analysis, which we call Deep Interval Neural Network (DINN). We perform experiments on an air pollution dataset with sensor uncertainty, and show that the DINN can produce accurate bounded estimates from uncertain input data. 
\end{abstract}

\section{Introduction}
\label{intro}

The use of deep neural networks (DNNs) is becoming more prevalent in important fields such as healthcare~\cite{ray2020ensemble}, physical sciences~\cite{raissi2018deep}, climate change~\cite{zantedeschi2019cumulo}, transportation~\cite{ jindal2018optimizing}, finance~\cite{heaton2017deep}, and the military---many of which include safety-critical applications. In such applications, the input data to DNNs is generally unstructured, perturbed, unlabeled, sparse or partially missing, and exposed to uncertainty from multiple sources. Within these applications, multiple complex problems are solved. Given the complexity of such problems, it is generally not the case that a single deep learning (DL) model provides the sole solution to a problem. In fact, these complex problems often involve a hierarchy of models composed of multiple computational models which solve different sub-problems that must share compatible inputs and outputs amongst them. For any predictive model within a complex problem, not quantifying the input uncertainty properly can lead to inaccurate and even disastrous decision-making~\cite{aoki2013comparative,zio2013industrial,wen2003uncertainty}. Therefore, given the high risks associated with uncertainty, it is desired to model the input uncertainty and to also obtain prediction uncertainty bounds from DNN models.

DNNs produce estimates that resemble frequentist estimators. Indeed, the parameters obtained are point estimates. Having point estimates for the parameters is a major drawback for uncertainty quantification because it does not allow to reason about the uncertainty in the parameters of the model and its corresponding prediction~\cite{murphy2012machine,hernandez2015probabilistic}. One solution would be using a Bayesian approach---e.g., Bayesian Neural Networks (BNNs). However, BNNs are more expensive to train than DNNs, have not been found to be superior predictors, nor gained wide adoption~\cite{wenzel2020good}. 

There has been recent work on deep learning (DL) algorithms that can quantify predictive model uncertainty in a Bayesian framework~\cite{hernandez2015probabilistic, gal2016dropout} that significantly improved the ability to obtain uncertainty estimates for DL model predictions. Nevertheless, quantifying the uncertainty in the input data in end-to-end training has been less studied within the DL community. It is common practice in some applications to add a small amount of noise to the training data to account for randomness and improve generalization~\cite{bishop1995training, czarnecki2013machine}. Others have integrated Bayesian methods with DNNs in order to account for input randomness~\cite{kendall2017uncertainties}. Nonetheless, more effort is required to train DNNs end-to-end in cases where the source of uncertainty is more than randomness in the environment. 

More specifically, there are two main types of uncertainty to quantify for predictive models. The first is called aleatory uncertainty and it arises due to randomness in the environment. Aleatory uncertainty is irreducible, as additional data does not guarantee certainty about the outcome. The second type is called epistemic uncertainty and it is due to incomplete knowledge in the environment or in the models. Epistemic uncertainty can be characterized by imperfect information, missing data, imprecision, incertitude, and model uncertainty~\cite{dalkey1967simulation, wen2003uncertainty, kreinovich2013computational,ferson2007experimental,beer2016we, der2009aleatory, walley1991statistical}. It is important to notice that in the general sense, epistemic \emph{model uncertainty} can be due to a model generating the data (e.g., a physics simulation model which generates the training data) and/or the predictive model itself (e.g., a DNN model)~\cite{ferson2007experimental,beer2016we}.\footnote{For DNNs in particular, model uncertainty is only due to the predictive model itself, while uncertainty around the input is called input uncertainty.} Above all, epistemic uncertainty can be reduced as we find more sources of information. Finally, let us highlight that both aleatory and epistemic uncertainty from different sources can be present in the input data.

Usually, the input data is obtained from measurements or observations which contain inherent aleatory and epistemic uncertainty. If the probability distributions of the observations can be assumed, then we can model them with random variables and use Bayesian or frequentist frameworks. However, oftentimes we do not have enough knowledge about the probabilistic nature of the observations---either because of incomplete knowledge or because they are impossible to obtain. Indeed, under such conditions of epistemic uncertainty, a strictly Bayesian or frequentist approach would require us to make probabilistic assumptions without enough knowledge, which does not guarantee reliable DL predictions. Examples of such cases in safety-critical applications are pandemics~\cite{macphail2010predictable} or failure modes of some materials~\cite{rao1997analysis}. To address the above challenges, interval analysis (IA) provides an alternative to model both types of uncertainty.

For this work, we assume that we do not possess enough knowledge about the epistemic uncertainties as to make probabilistic assumptions for the data, but that we only possess knowledge of the uncertainty bounds of it (in the form of upper and lower bounds). As a result, we use IA to model the input data epistemic uncertainty\footnote{Epistemic uncertainty expressed with interval numbers is also called incertitude~\cite{ferson2007experimental, calder2018quantification}.}. In this paper we describe the development of a novel deep interval-valued neural network (DINN) to quantify epistemic uncertainty in the input and model. The main goal of this work is to quantify uncertainty of input data and propagate it through a deep neural network via IA in order to obtain the corresponding parameter and prediction uncertainty. Crucially, with IA algorithms, the challenge is to reduce the overestimation of prediction uncertainty as much as possible. This work can be used broadly for applications in reinforcement learning, AI safety, data privacy, and DNNs verification.

In this paper we seek to present our work in a self-contained way that is easy to follow for machine learning researchers with expertise in DNNs but not familiar with IA, as well as machine learning researchers with expertise in IA but not well-versed with the inner-workings of DNNs. This paper is organized as follows: Sec.~\ref{relatedwork} presents related work, Sec.~\ref{background} presents the fundamentals of IA computations relevant to this work. In Sec.~\ref{DINNsup} we present in detail the development the DINN, some of which are extensions of DNNs with modifications for interval computations. In Sec.~\ref{sec:experiments} we present experiments with an Air quality dataset and compare our method with a Bayesian approach.  
\section{Related Work}
\label{relatedwork}

For DNNs, predictive model uncertainty quantification has customarily taken place by using Bayesian Neural Networks (BNNs) or Bayesian approximations. In \cite{hernandez2015probabilistic}, the authors present Probabilistic Backpropagation, a Bayesian neural network method to train deep neural networks and obtain parameter uncertainty. Shortly after, \cite{gal2016dropout} developed a method called Monte Carlo Dropout, which averages Monte Carlo simulations of predictions using Dropout~\cite{srivastava2014dropout} at test time in order to obtain a predictive uncertainty estimate. In deep reinforcement learning, state uncertainty in Partially Observable Markov Decision Processes (POMDPs) was studied by~\cite{hausknecht2015deep} using Long Short-Term Memory (LSTM) units in order to \emph{remember} past states and has become an important artifact for this purpose in reinforcement learning.

On the other hand, training DNNs end-to-end with uncertain input data has not received as much attention. Most of the work adds noise to the data with known probability distributions~\cite{czarnecki2013machine} or uses sampling methods~\cite{mcdermott2019bayesian}. In many safety-critical applications, we do not have enough information to select a probability distribution for the input, as in conditions of epistemic uncertainty, but can gain access to bounds on the data by expert knowledge or by the nature of the data acquisition~\cite{ferson2007experimental}. In such cases, when only the bounds of the input data are known, IA~\cite{moore1979methods} offers an alternative to model the uncertainty with guarantees for the prediction. 

More generally, IA algorithms are used for rigorous error bounds, result verification, sensitivity analysis, and uncertainty quantification~\cite{moore1979methods, neumaier1990interval}. IA has been studied since the late 1950s~\cite{moore1959interval} and has found a recent resurgence to solve multiple important problems. For example, methods have been developed to solve systems of interval equations~\cite{moore1979methods, neumaier1990interval}, optimization with intervals~\cite{hansen2003global, fiedler2006linear}, and more recently for numerical integration with arbitrary-precision arithmetic~\cite{johansson2017arb, johansson2018numerical}. For uncertainty quantification in particular, IA has been used for inference with imprecise probabilities~\cite{walley1991statistical}, matrix inequalities~\cite{ben2002tractable}, partial differential equations with finite element methods~\cite{muhanna2001uncertainty}, and data processing~\cite{kreinovich2013computational}.  

Recently, IA has been used as a surrogate to verify DNNs but not in a way that can be directly used for training~\cite{liu2019algorithms, adam2016bounding}. Symbolic IA has also been used for security analysis of DNNs~\cite{wang2018formal}, but not with mathematically rigorous IA algorithms as used for the other computational applications previously mentioned.  

For neural networks (NNs) using AI for training and inference, work that began in the 1990s used interval-valued inputs to train shallow NNs for smaller datasets but, with few exceptions, not trained using gradient-based optimization. Nonetheless, the developments were valuable to demonstrate the concept of learning with interval-valued inputs. For example,~\cite{ishibuchi1991extension, hernandez1993interval} developed Backpropagation for interval arithmetic using a a fixed-size feed-forward two-layer neural network, \cite{freitag2011recurrent} developed an algorithm to train interval recurrent neural networks, \cite{freitag2012particle} trained a recurrent neural network using particle swarm optimization. Other work, has attempted to train neural networks with interval values but without using IA, which makes it is less rigorous~\cite{yang2019interval}. In fact, work in this fashion carried on until recently, using older neural network architectures and optimization methods~\cite{yang2012smoothing, yang2018l1, yang2019interval}. Remarkably, first-order gradient-based optimization methods for interval inputs have not been widely studied in the literature. In contrast, other optimization methods for interval input have been studied~\cite{neumaier1990interval,moore1979methods, hansen2003global}, which include linear programming, convex programming, Newton methods, and global optimization. This is perhaps because in numerical analysis at-large, first-order methods have been less preferred than the others mentioned. However, first-order methods found great success and appeal in DNNs. 

\paragraph{Challenges with IA algorithms}
First, IA algorithms that do not use \textit{rigorous} IA (as in~\cite{moore1979methods, revol2017introduction}) do not guarantee enclosure of the real-valued model solution. Not having rigorous IA in place defeats the purpose of reliable computing. Second, na\"ive IA algorithms suffer from a setback called \emph{interval dependency}. When interval dependency is not reduced, it leads to significant overestimation of the interval bounds in the computations. Consequently, designing novel IA algorithms that reduce interval dependency has been the bulk of the work in IA for the last twenty years. With this in mind, our contributions are as follows: 
    (\emph{i}) Present a novel deep learning algorithm (DINN) able to quantify and propagate input and parameter epistemic uncertainty via IA and produce bounded guaranteed predictions that are reasonably sharp. (\emph{ii}) From an optimization perspective, our contributions is to study the behavior of first-order gradient-based methods with IA. 

\section{Interval Analysis}
\label{background}

In this section we present the elements for the numerical computations using interval input for DNNs. In particular, we first introduce the use of monotonic interval functions and the interval Lipschitz property in order to achieve interval bounds of the solution that are as sharp as possible. We then explain in detail how the the interval gradient-based optimization for the DINN was developed. For readers new to IA, more background is included in the appendix.

\subsection{Interval Arithmetic}\label{intarith}
Let $\mathbb{IR}$ be the set of interval real numbers. Interval real numbers are a closed and bounded set of real numbers~\cite{moore1979methods}, such that if $\Vec{X} \in \mathbb{IR} $ is an interval, its endpoints are $\underline{X} = \inf(\Vec{X}) \in \mathbb{R}$ and $\overline{X}= \sup(\Vec{X}) \in \mathbb{R}$. \textit{Henceforth, boldface is used to represent interval scalars, matrices, and vectors}. The basic arithmetic operations with real intervals are different than with real numbers and are introduced in~\cite{moore1979methods,moore2009introduction}. Any of the four elementary interval arithmetic operations between two intervals can be defined as the following set

\begin{equation}\label{genarithm}
    \Vec{X} \diamond \Vec{Y} = \{x \diamond y : x \in \Vec{X}, y \in \Vec{Y}  \},
\end{equation}
where $\diamond$ can be the operator for addition, subtraction, multiplication, or division. 

\paragraph{Describing Interval Numbers}
An interval number $\vec{x} \in \mathbb{IR}$ can be described by its infimum and supremum (i.e., its endpoints), as $\vec{x} = [\underline{x}, \overline{x}] = \{x \in \mathbb{R}:\underline{x} \leq x \leq \overline{x}  \}$.

The same interval number $\vec{x} \in \mathbb{IR}$ can be described by its midpoint $\texttt{mid}(\vec{x})  = \frac{1}{2}(\underline{x} + \overline{x})$, radius $\texttt{rad}(\vec{x}) = \frac{1}{2}(\overline{x}-\underline{x})$, width $\texttt{w}(\vec{x}) = 2\cdot \texttt{rad}(\vec{x})$, and \emph{uncertainty level} $\beta$. The \emph{uncertainty level} of an interval $\vec{x}$ is measured in terms of percentage as the width of the interval $\boldsymbol{\beta}$, as 

\begin{equation}
    \beta = \texttt{w}(\boldsymbol{\beta}),
\end{equation}  
where 
\begin{equation}
    \boldsymbol{\beta} = \left[-\frac{\texttt{rad}(\vec{x})}{\texttt{mid}(\vec{x})},\frac{\texttt{rad}(\vec{x})}{\texttt{mid}(\vec{x})} \right].
\end{equation}
With these measures in place, $\vec{x} \in \mathbb{IR}$ can described as

\begin{equation}\label{uncertlevel}
    \vec{x} = \texttt{mid}(\vec{x}) (1+ \boldsymbol{\beta}) = \texttt{mid}(\vec{x}) \Big(1- \Big| \frac{\texttt{rad}(\vec{x})}{\texttt{mid}(\vec{x})} \Big| , 1+ \Big| \frac{\texttt{rad}(\vec{x})}{\texttt{mid}(\vec{x})} \Big| \Big), 
\end{equation}
which allows to conveniently describe the interval in terms of its midpoint value and uncertainty level.
Finally, an interval number $\vec{x} \in \mathbb{IR}$ is said to be degenerate if its radius is zero, i.e., $\underline{x}=\overline{x}$. If $\vec{x}$ is a degenerate interval, then it is a singleton with $x \in  \mathbb{R}$.

\subsection{Interval Functions}\label{intfunc}
The image of the sets $\Vec{X}_1, \cdots, \Vec{X}_n$ under a real-valued function $f$ is \\ $f(\Vec{X}_1, \dots , \Vec{X}_n) = f(x_1, \dots , x_n),~\text{for}~ x_1 \in \Vec{X}_1, \dots, x_n \in \Vec{X}_n$. Furthermore, an interval-valued function $F: \mathbb{IR}^n \mapsto \mathbb{IR}$ is an interval extension of a real-valued function $f: \mathbb{R}^n \mapsto \mathbb{R}$ if for degenerate interval arguments, $F$ coincides with $f$ such that

\begin{equation}
F([\underline{X},\overline{X}]) = F([x,x]) = f([x,x]), ~\quad \text{for all}~ x \in \mathbb{R}.
\end{equation}

For non-degenerate intervals, real functions $f$ using real arithmetic give rise to different results for the range of interval extensions $F$.  

\paragraph{Isotonic Inclusion of Interval Functions}
Isotonic inclusion is a property of IA that guarantees that an interval extension function $F$ contains the exact upper and lower bounds of a real-valued function $f$. 

\begin{theorem}\label{fundamentalthm}
If $F$ is an inclusion isotonic interval extension of $f$, then for interval arguments $\Vec{X}_i$

$f(\Vec{X}_1, \dots , \Vec{X}_n)  \subseteq F(\Vec{X}_1, \dots , \Vec{X}_n) .$

\end{theorem}

While isotonic inclusion guarantees enclosure of a function, the resulting bounds could become unrealistically wide due to \textit{interval dependency}. Consequently, a fundamental goal of IA algorithms is to reduce interval dependency to obtain the sharpest possible bounds in the range of a function.

\paragraph{Interval Dependency}
Overestimation due to interval dependency occurs when the same interval variables appear more than once in an expression. Such dependency is due to the memoryless structure of interval arithmetic~\cite{neumaier1990interval}. In general, because interval arithmetic does not posses distributive, additive, and multiplicative inverse properties, real functions using real arithmetic might not produce the same results as their interval extensions using interval arithmetic~\cite{moore2009introduction}. 

\paragraph{Monotonic Interval Functions}
Interval extensions of monotonic real functions produce sharp bounds that general interval extensions might not be able to produce. Therefore, using the monotonic property of interval functions allows us to easily compute the sharpest enclosure.

\paragraph{Probabilistic Interpretation}
Interval solutions (e.g., predictions) guarantee to enclose all possible solutions associated with different probabilistic distributions of the input data, either symmetrical
or not.~\cite{zhang2010interval}. It is important to note that interval numbers are not the same as uniform probability distributions. Additionally, they are not to be confused with confidence intervals. See the appendix for comparisons.

\subsection{Interval Lipschitzness}\label{intlip}
Natural real extensions of real rational functions satisfy a Lipschitz condition with a constant $L$. With the Lipchitz property, we guarantee that small changes to the input result in small changes to the output by a constant $L$, when using interval functions or solving interval systems of equations. 

\begin{definition}
An interval extension $F$ is Lipschitz in $\Vec{X}_0$ if there is a constant $L$ s.t. $\texttt{w}(F(\Vec{X})) \leq L \texttt{w}(\Vec{X})$ for every $\Vec{X} \subseteq \Vec{X}_0$, 
\end{definition}
where $\textit{\texttt{w}}$ measures the width of the interval.

\begin{lemma}
If a real-valued function $f(x)$ satisfies an ordinary Lipschitz condition in $\Vec{X}_0$, 

\begin{equation}
    |f(x) - f(y)| \leq L |x-y|,~\quad \text{for}~x,y \in \Vec{X}_0.
\end{equation}
then the set image of $f$ is a Lipschitz interval extension in $\Vec{X}_0$.
\end{lemma}
Proof in \cite{moore2009introduction}.

Moreover, the Lipschitz property of intervals is important because compositions of interval functions such as $H(\Vec{X}) = F(G(\Vec{X}))$ for inclusion isotonic interval functions $F$ and $G$ are Lipschitz and inclusion isotonic if the individual functions are also Lipschitz, which reduces overestimation.  

\section{Deep Interval Neural Network}\label{DINNsup}
This section will explain in detail the development of the The Deep Interval Neural Network (DINN) and generalize definitions of supervised learning to interval values. Detailed calculations for the interval gradient derivations are provided in the appendix.  
The DINN is the interval extension of a real-valued DNN (see Sec.~\ref{intfunc}) to process interval-valued matrices of any dimensionality (i.e., interval tensors). The DINN presented in this paper is a supervised learning algorithm in a regression setting which seeks to learn an interval predictive model $F: [\underline{X},\overline{X} ] \rightarrow [\underline{Y},\overline{Y} ]$  $ \in \mathbb{IR}$. In order to achieve this goal, the DINN is trained using the $\Vec{X}$ interval-valued features in a $d$-dimensional space with $n$ samples, along with its known $\Vec{Y}$ interval-valued targets. This composes the training set of $n$ samples $\mathcal{T} = \{(\Vec{X}_1, \Vec{Y}_1), \dots ,(\Vec{X}_n, \Vec{Y}_n) \} $ where $\Vec{X} \in \mathbb{IR}^{n \times d}$ is the feature space and $\Vec{Y}\in \mathbb{IR}^n$ is the target space. As its output, the DINN computes for each sample $i$ an estimate of the interval target $\widehat{F}(\Vec{X}_i)$. In order to do so, the training algorithm iteratively reduces the difference between the true known target $\Vec{Y}_i$ and its prediction target $\widehat{F}(\Vec{X}_{i})$ by minimizing a loss function $\mathcal{L}(\widehat{F}(\Vec{X}_{i}), \Vec{Y}_i;\Vec{W})$, such as mean squared error, parameterized by $\Vec{W}$. 

A na\"ive implementation of algorithms based on rigorous interval arithmetic usually leads to overestimation of bounds because of interval dependency (See Section~\ref{intfunc}). On the other hand, not employing rigorous interval arithmetic does not guarantee enclosure by the isotonic inclusion property (See Theorem \ref{fundamentalthm}). Thus, the core of the work in the DINN consists of algorithmic design using rigorous interval arithmetic to achieve guaranteed and sharper bounds of the model prediction. The following sections delve into the details of the DINN.  

\subsection{Fully Connected DINN}
\begin{wrapfigure}{l}{0.5\textwidth}
	\includegraphics[width=\linewidth]{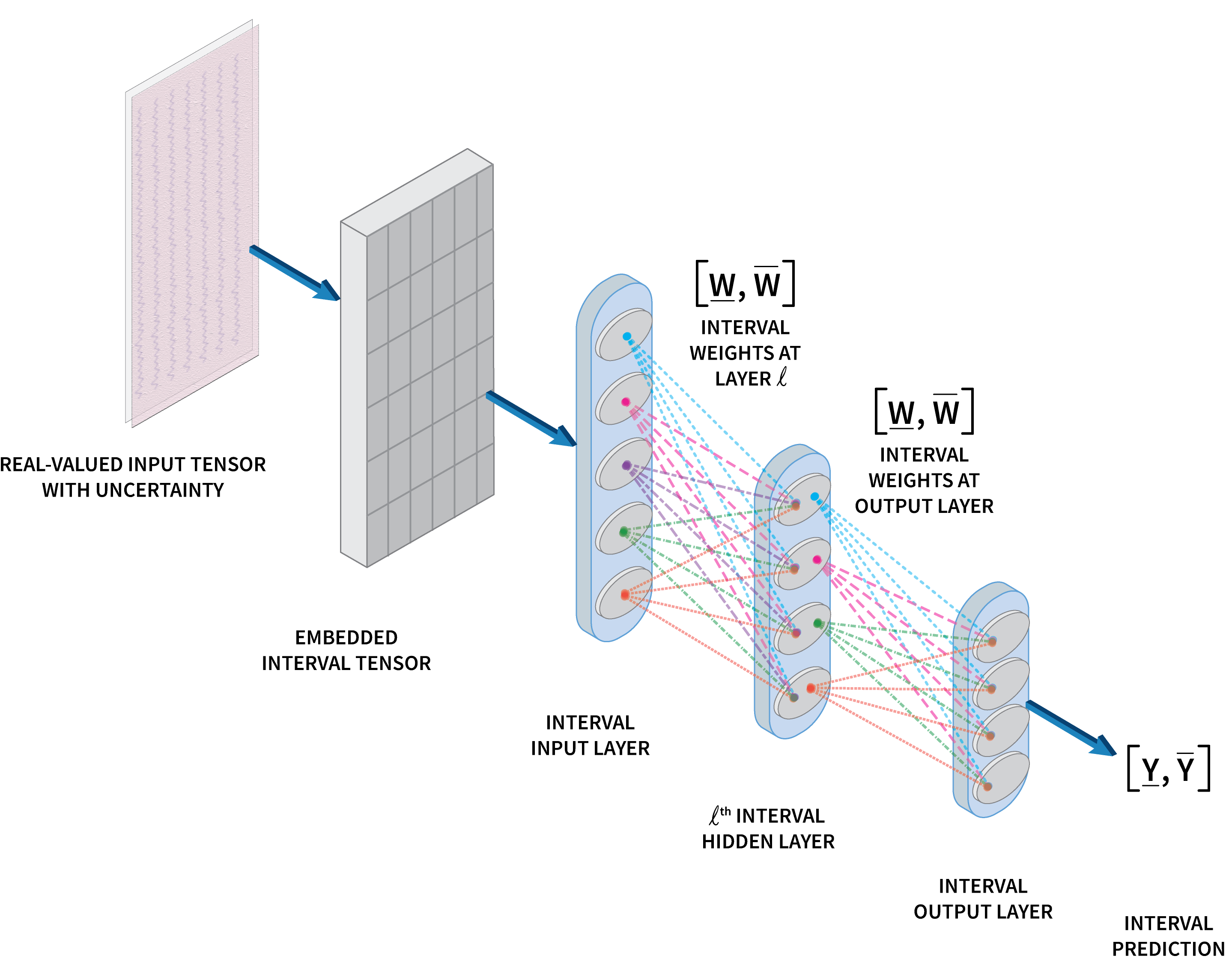}
	\caption{\label{fig:dinn_arch} DINN architecture. The input, parameters, and output are intervals.\vspace{-0.1mm}}
\end{wrapfigure}

For the DINN and DNNs in general, the predictive interval function $\widehat{F}(\Vec{X})$ at the output layer $L$ is composed of the functions of the hidden layers, such that $\widehat{F}(\Vec{X}) = F^{(L)} \big( \dots F^{(\ell)}(\dots F(\Vec{X})) \big)$, for $\ell=1 \dots L$ where $\ell$ is the layer number. A depiction of the DINN is shown on Fig.~\ref{fig:dinn_arch}.

Hence, the DINN is a composition of interval-valued functions where each of these functions is one the $\ell$ layers of the network after passing through a nonlinear activation function. A crucial aspect of the development of the DINN is requiring the nonlinear activation function at each layer $F^{(\ell)}$ be monotonic and Lipschitz so that the final output can also be Lipschitz and as sharp as possible (see Sec.~\ref{intfunc} and Sec.~\ref{intlip}). Moreover, the computations of the algorithm to obtain the gradients of the interval functions are ordered in a way to reduce interval dependency by eliminating repeated values of the same variable. With these elements in place, the interval predictions $\widehat{F}(\Vec{X})$ of the DINN can be as sharp as possible.    

In the regression setting, for the final layer, the prediction is simply the affine transformation of final output without a \texttt{ReLU}. Finally, the loss at the end of each epoch is computed using Mean Squared Error (MSE) for each mini-batch of size $B$, as $\mathcal{L}(\widehat{F}(\Vec{X}), \Vec{Y}) = \frac{1}{2B} \sum_{i=1}^{B}(\widehat{F}(\Vec{X}_{i}) - \Vec{Y}_i)^2$.

The loss is an interval scalar. Notice that \texttt{ReLU} is a monotonically increasing function, which allows for calculation of sharp interval enclosures. The \texttt{ReLU} is the standard activation function for the DINN but it can be replaced by a different activation function with ease.
\paragraph{Interval Computations}
To our knowledge, there is no open-source Python-compatible library for rigorous interval matrix computations that follow the IEEE 1788-2015 standard~\cite{revol2017introduction}. Thus, all matrix computations are performed with the INTLAB~\cite{rump1999intlab} toolbox in MATLAB which has been thoroughly benchmarked. The raw real-valued features $X$ and targets $Y$ in the training set $\mathcal{T}$ are converted as needed to intervals $\{ \Vec{X}_i,\Vec{Y}_i \}_{i=1}^{n}$ in an \emph{embedding} layer, according to their upper and lower bounds. We use Eq.~\ref{uncertlevel} in order to define the interval input for any given subset of input data under uncertainty. Test-time features are also assumed to undergo a pre-processing step to be converted to intervals.   

\paragraph{Training}\label{training}
During training, we find a matrix of optimal parameters $\Vec{W} \in \mathbb{IR}$ for each layer of the network by minimizing the loss function. In order to minimize the interval loss function $\mathcal{L}( \widehat{F}(\Vec{X}_{i}), \Vec{Y}_{i})$, we use first-order gradient-based optimization. In particular, we use mini-batch stochastic gradient descent (SGD) and some of its variants; SGD with momentum~\cite{polyak1964some} and Adam~\cite{kingma2014adam}. The vanilla gradient descent update rule at step $k+1$ for sample $i$ at layer $\ell$ is defined as

\begin{equation}\label{gradstep}
\Vec{W}_{k+1}^{(\ell)} = \Vec{W}_k^{(\ell)} - \alpha_k \nabla_{\Vec{W}_k}^{(\ell)} \mathcal{L}( \widehat{F}(\Vec{X}_i), \Vec{Y}_i),  
\end{equation}
where at each step $k$, $\Vec{W}_{k}$ is the interval weight matrix, $\alpha_k$ is the step size, and $\nabla_{\Vec{W}_k} \mathcal{L}( \cdot)$ are the gradients of the loss function with respect to parameters $\Vec{W}_{k}$.

\paragraph{First-order gradient optimization with IA}
Training real-valued DNNs with SGD presents numerical problems like exploding and vanishing gradients. In the case of the DINN, the interval input makes these numerical problems more pronounced. Furthermore, implementing SGD and its variants with heuristics made for real-valued DNNs did not instantly led to good results in the DINN. Since the gradients are interval-valued, they are more likely to have large cliffs and explode.  

In order to avoid these problems, the implementation of the DINN takes into consideration three important factors for optimization. First, we required that the interval functions in the DINN are Lipschitz continuous (see~\ref{intlip}). Second, we design the algorithms in order to reduce interval dependency. Third, because of interval matrix multiplications, we want to avoid large weights than can lead to exploding gradients. This is avoided by:

\begin{itemize}
    \item Initializing the weights to smaller random values than for DNNs (usually a factor of magnitude smaller).
    \item Alternatively to initializing with small random values, pretraining an equivalent DNN with real-valued input to provide a warm start.
    \item Clipping the gradients at every iteration.
    \item Annealing the learning rate or having a learning rate decay schedule.
\end{itemize}

Both vanilla SGD and SGD with momentum do not require changes to the update rule for the DINN. However, the Adam optimization algorithm does require changes. These changes give rise to Interval Adam (I-Adam), which is presented on Algorithm~\ref{IvanIntval}. Notice that in I-Adam, the second-order biased estimates $v_k$ are modified so that they are real-valued---by squaring only the midpoints of the interval gradients in $\texttt{mid}(\mathcal{\Vec{G}}_k)^2$. 

\begin{algorithm}[ht]
	\caption{I-Adam: Adam Algorithm for Interval Input}\label{IvanIntval}
		\KwIn{Step size $\alpha$  } 
        \KwIn{Decay rates $\beta_1, \beta_2 \in [0,1)$ } 		
		\KwIn{Initial parameters $\Vec{W} \in \mathbb{IR}$ } 
		Initialize $\Vec{m}_0 \in \mathbb{IR} = 0$ \\
		Initialize $v_0 \in \mathbb{R} = 0$ \\
		Initialize $k = 0$  \\
		\While{stopping criteria not met}{
		$k \leftarrow k + 1$ \\
		$\Vec{G}_k \leftarrow \nabla_{\Vec{W}}\mathcal{L}_k$ compute interval gradients at step $k$ \\
		$\Vec{m}_k \leftarrow \beta_1 \Vec{m}_{k-1} + (1-\beta_1) \Vec{G}_k$ biased interval first moment estimate (interval)\\
		$v_k \leftarrow \beta_2 v_{k-1} + (1-\beta_2) \cdot \texttt{mid}( \Vec{G}_k)^2$ biased real second moment estimate (real)\\
		$\widehat{\Vec{m}}_k \leftarrow \Vec{m}_k / (1- \beta_1^k) $\\
		$\widehat{v}_k \leftarrow v_k/(1-\beta_2^k) $\\
		$\Vec{W}_k \leftarrow \Vec{W}_{k-1} - \alpha \cdot \widehat{\Vec{m}}_k / (\sqrt{\widehat{v}_k}+\epsilon)$
		
		}
		\Return{$\Vec{W}_k$}
\end{algorithm}

\paragraph{Regularization}
Regularization is used for two purposes in the DINN. First, regularization is used to prevent overfitting to the training set, with the goal of achieving better performance in unseen data. Second, it stabilizes the solution and narrows the enclosure bounds---which for finding interval gradients is of crucial importance. Two forms of regularization are used for the DINN. $L1$ regularization uses the $l_1$  norm, which is $\Omega(\Vec{W}) = \|\Vec{W}\|_1$ and $L2$ regularization which is  defined as $\Omega(\Vec{W}) = \sum_{m}\sum_{n}\Vec{W}_{m,n}^2$.

\section{Experiments}\label{sec:experiments}
After an extensive search we found that large interval datasets are not yet openly available in machine learning repositories. Thus, we converted a real-valued benchmarked dataset into an interval-valued dataset by using uncertainty quantification of the input, thereby assuming expert knowledge. In general, a real-valued dataset can be converted to interval-valued if we possess knowledge on the uncertainty bounds.  The main goal of the experiments is to demonstrate that we can propagate uncertainty through the DINN and obtain a guaranteed bound of the equivalent real-valued solution.  

\paragraph{Air Quality dataset}
The air quality dataset consists of one-year sensor measurements for common air pollutants obtained from the UCI repository and developed in~\cite{de2008field}, with follow-up papers in~\cite{de2009co, de2012semi}. The set-up consists of five Metal Oxides sensors (measuring electrical resistance) and environmental measurements (temperature, humidity) in a polluted urban environment in an Italian city. In parallel, a colocated certified analyzer measured the ground truth concentrations for the pollutants. When the dataset was presented, it was reported that the data contained sensor drift~\cite{de2008field} and that sensor measurements became erroneous over time. This is common for many sensors, and the uncertainty due to drift cannot be easily assigned a probability distribution to account for noise, while recalibration is expensive. 

In previous studies of this dataset~\cite{de2008field, de2009co, de2012semi}, different techniques were performed to ignore large blocks of data exposed to significant sensor drift and other errors. Some of the techniques seem to benefit from domain expertise and hindsight. While that could be acceptable in some settings, we recognize that more often, there are urgent time limitations to make predictions in safety applications. Indeed, this raises a more general question as to how to use DL with uncertain sensor data streams for safety applications. Thus, our general motivation with the DINN is develop a workhorse that can be used to make predictions with input data under uncertainty, without claiming to solve the problem of data acquisition under uncertainty. In this experiment, we account for the uncertainty of the sensor drift, environmental factors, and other unknowns in a pre-processing step by converting the real-valued input to intervals using assumed expert knowledge. 
\paragraph{Epistemic Uncertainty in the Input}
Uncertainty quantification (UQ) of the input (i.e., generating the interval input embedding layer) is largely a problem-specific task and thus not the direct goal of this work. Nonetheless, we perform UQ of the input to demonstrate the overall process of converting real-valued measurements into intervals before training the DINN.  
 \begin{wrapfigure}{r}{0.4\textwidth}
	\vspace{-0.1mm}
	\includegraphics[width=.9\linewidth]{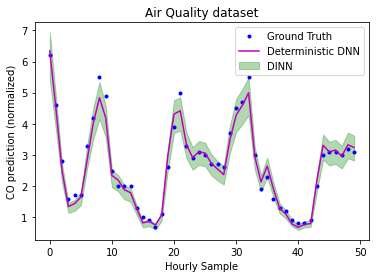}
	\caption{\label{fig:COpred} Test CO Predictions during a week in spring-time. 84\% of the ground truth test set samples were covered by the interval prediction.\vspace{-0.1mm}}
\end{wrapfigure}
In~\cite{de2008field}, it is noted that spring time sensor measurements contain significant uncertainty due to sensor drift and pollutant concentrations~\cite{de2008field}. Nitrogen oxides (NOx) measurements were particularly more problematic. The complex interactions of known and unknown factors that contribute to sensor drift and changes in pollutant concentrations would make it unrealistic to build a sound probabilistic measurement model. As we can see that the input uncertainty is greatly in part epistemic, intervals will be chosen to model the uncertainty. Using intervals, we can encode expert knowledge in the input without selecting a probability distribution. From the error between ground truth measurements of NOx and sensor measurements reported in~\cite{de2008field}, we can use a safe uncertainty level of $ \beta = 10\%$ for this feature for measurements between March to June (i.e., an interval radius of 5\% for each sample, See Eq.~\ref{uncertlevel}). 
We then used an uncertainty level of $\beta=1\%$ for all other features in $\Vec{X}$ and \emph{degenerate} intervals for the targets $\Vec{Y}$ (i.e., the targets are $Y \in \mathbb{R}$). This set-up implies that we have uncertainty in the observations but not in the targets.

The DINN is flexible to different approaches to model the interval uncertainty of the input---either by using more informed expert bounds in selected features in $\Vec{X}$, with the aid of Monte Carlo sampling to obtain bounds, and by including uncertainty in the targets $\Vec{Y}$.  

\paragraph{Model Details}
We trained the DINN with SGD, SGD with momentum, Adam without modifications, and I-Adam (Algorithm~\ref{IvanIntval}). We use three hidden layers, 500 units per layer, and 100 epochs. We use randomized search cross-validation~\cite{bergstra2012random} with 20 folds to find optimal hyperparameters for the DINN and keep hours in the same week with the same selection index. We do not take into consideration time dependencies and thus ignore the time dimension (as in~\cite{de2012semi}). We use the following seven features: \textbf{\texttt{PT08.S2(NMHC), PT08.S3(NOx), PT08.S4(NO2), PT08.S5(O3), T, RH, AH}}  and the target is \textbf{\texttt{CO}} from the certified sensor. 

\paragraph{Comparison}
We compare with a Bayesian approximation approach. The pitfall of this approach for the current dataset is that we have to select probability distributions without having enough knowledge about the uncertainty of the inputs. To account for the input uncertainty, we added noise $\varepsilon$ to the normalized features using a Gaussian distribution with a variance equal to the interval radius, $\varepsilon \sim  \mathcal{N}(0,\,0.05)$.  Then, to account for \emph{model uncertainty} we used Monte Carlo Dropout~\cite{gal2016dropout} using the jittered input with 0.5 dropout probability after every hidden layer. Training with noise can improve generalization performance and is equivalent to Tikhonov Regularization~\cite{bishop1995training}. Besides regularization, training with noise has also been used to account for randomness~\cite{holmstrom1992using, czarnecki2013machine}. Finally, we also train a model with the same settings with Monte Carlo Dropout but without jitter.   

\begin{figure}[ht]
    \centering
    \begin{minipage}{0.45\textwidth}
        \centering
        \includegraphics[width=0.9\textwidth]{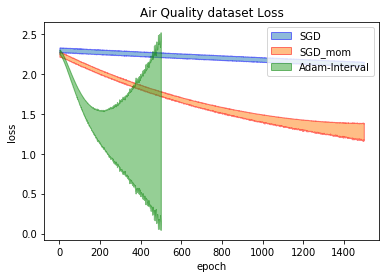} 
        \caption{\label{fig:300iters}Training loss for optimizers up to 1400 epochs. Interval Adam performs better early on and then blows up.}
    \end{minipage}\hfill
    \begin{minipage}{0.45\textwidth}
        \centering
        \includegraphics[width=0.9\textwidth]{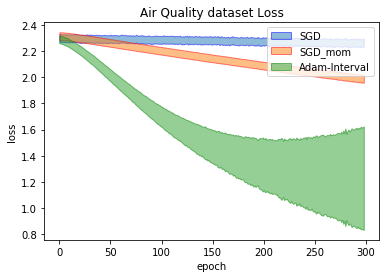} 
        \caption{\label{fig:1400iters}Training loss for optimizers up to 300 epochs }
    \end{minipage}
\end{figure}

\paragraph{Results}
Fig~\ref{fig:COpred} shows that the predictions of the DINN are not too wide and enclose the predictions of the underlying real-valued DNN counterpart. Furthermore, the interval prediction encloses 84\% of the ground truth test set samples. Table~\ref{tab:mainres} shows the average interval mean squared error (MSE) for the various optimization algorithms examined. We show that interval uncertainty in the input data always results in interval uncertainty in the prediction. Also, as shown on Figs~\ref{fig:300iters} and \ref{fig:1400iters}, I-Adam gives the best performance but it quickly blows up if not given a performance-based stopping criterion. Adam without modifications blows up only after two iterations (not shown). Interestingly, the jittered input did not make a noticeable difference using Monte Carlo Dropout (Table ~\ref{tab:mainres}). It appears that the jitter only has a small regularizing effect but did not add to the predictive uncertainty.

\begin{table*}[ht]
\caption{Average interval MSE test set performance under different optimizers for the DINN and Comparision with MC Dropout. MC Dropout uses Adam. }
\label{tab:mainres}
\begin{tabular}{@{}lllccccc@{}}
\toprule
 &  &  & \multicolumn{3}{c}{Average Interval MSE for DINN} & \multicolumn{2}{c}{Avg. MSE $\pm$ std for MC Dropout} \\ \midrule
\textbf{Dataset} & \multicolumn{1}{c}{\textbf{n}} & \multicolumn{1}{c}{\textbf{d}} & \textbf{SGD} & \textbf{SGD-Mom.} & \textbf{Interval Adam} & \textbf{MC Dropout+Jit.} & \textbf{MC Dropout} \\
Air Quality & 7344 & 7 & {[}0, 0.24{]} & {[}0.09, 0.33{]} & {[}0.1, 0.31{]} & 0.21 $\pm$ 0.07 & 0.21 $\pm$ 0.07 \\ \bottomrule
\end{tabular}
\end{table*}

\section{Conclusions and Future Work}
We have demonstrated that we can quantify and propagate uncertainty in the input data and obtain the corresponding prediction uncertainty by using the developed DINN algorithm. We also show that the uncertainty estimates of the DINN are realistic and not overly wide. 

This work is only the beginning of the great potential to use IA with DNNs. The DINN needs to be extended to include recurrent and convolutional layers. Additionally, the integration with Bayesian methods is also promising, to include cases when probabilistic knowledge about the uncertainties is available and the algorithm can actively interact in the environment to update its belief.

\bibliography{bibliography}
\bibliographystyle{plain}


\end{document}